\documentclass[preprint,12pt]{elsarticle}

\usepackage{graphicx}
\usepackage{amssymb}
\usepackage{amsthm}
\usepackage{amsmath}
\usepackage{lineno}
\usepackage{xcolor}
\usepackage{url}
\usepackage{xparse}
\usepackage{algpseudocode,algorithm,algorithmicx}
\usepackage{booktabs}

\usepackage{footnote}
\makesavenoteenv{tabular}
\makesavenoteenv{table}

\journal{Chaos, Solitons \& Fractals}

\begin{document}

\definecolor{violet}{RGB}{153,51,255}
\definecolor{maroon}{RGB}{153,76,0}

\newcommand\PC[1]{{\textcolor{red}{[\textbf{PC: #1}]}}}
\newcommand\RS[1]{{\textcolor{violet}{[\textbf{RS: #1}]}}}
\newcommand\MA[1]{{\textcolor{blue}{[\textbf{MA: #1}]}}}
\newcommand\FG[1]{{\textcolor{cyan}{[\textbf{FG: #1}]}}}
\newcommand\HL[1]{{\textcolor{maroon}{[\textbf{HL: #1}]}}}
\newcommand\PB[1]{{\textcolor{green}{[\textbf{PB: #1}]}}}

\begin{frontmatter}

\title{COVID-ABS: An Agent-Based Model of COVID-19 Epidemic to Simulate Health and Economic Effects of Social Distancing Interventions}

\author[cidic,ifnmg,minds]{Petr\^onio C. L. Silva}
\author[cidic,ifnmg]{Paulo V. C. Batista}
\author[ifnmg]{H\'elder S. Lima}
\author[minds,ppgee]{Marcos A. Alves}
\author[minds,ufmg]{Frederico G. Guimar\~aes }
\author[minds,ufop]{Rodrigo C. P. Silva}

\address[cidic]{Grupo de Pesquisa em Ci\^encia de Dados e Intelig\^encia Computacional - $\{ci\partial ic\}$}
\address[ifnmg]{Instituto Federal do Norte de Minas Gerais (IFNMG), Brazil}
\address[minds]{Machine Intelligence and Data Science (MINDS) Laboratory, Federal University of Minas Gerais, Brazil}
\address[ufmg]{Department of Electrical Engineering, Universidade Federal de Minas Gerais (UFMG), Brazil}
\address[ppgee]{Graduate Program in Electrical Engineering - Universidade Federal de Minas Gerais - Av. Ant\^onio Carlos 6627, 31270-901, Belo Horizonte, MG, Brazil}
\address[ufop]{Department of Computer Science, Universidade Federal de Ouro Preto (UFOP), Brazil}

\begin{abstract}
The COVID-19 pandemic due to the SARS-CoV-2 coronavirus has directly impacted the public health and economy worldwide. To overcome this problem, countries have adopted different policies and non-pharmaceutical interventions for controlling the spread of the virus. This paper proposes the COVID-ABS, a new SEIR (Susceptible-Exposed-Infected-Recovered)  agent-based model that aims to simulate the pandemic dynamics using a society of agents emulating people, business and government. Seven different scenarios of social distancing interventions were analyzed, with varying epidemiological and economic effects: (1) do nothing, (2) lockdown, (3) conditional lockdown, (4) vertical isolation, (5) partial isolation, (6) use of face masks, and (7) use of face masks together with 50\% of adhesion to social isolation. In the impossibility of implementing scenarios with lockdown, which present the lowest number of deaths and highest impact on the economy, scenarios combining the use of face masks and partial isolation can be the more realistic for implementation in terms of social cooperation. The COVID-ABS model was implemented in Python programming language, with source code publicly available. The model can be easily extended to other societies by changing the input parameters, as well as allowing the creation of a multitude of other scenarios. Therefore, it is a useful tool to assist politicians and health authorities to plan their actions against the COVID-19 epidemic.


\end{abstract}

\begin{keyword}
COVID-19 \sep Agent-based Simulation \sep epidemic models \sep SEIR


\end{keyword}

\end{frontmatter}



\section{Introduction}
\label{S:1}


The Coronavirus disease 2019 (COVID-19) pandemic is an ongoing outbreak, caused by severe acute respiratory syndrome coronavirus 2 (so-called SARS‑CoV‑2). The outbreak was identified in Wuhan, China, in December 2019 \cite{Huang2020}. The World Health Organization (WHO) declared the outbreak a Public Health Emergency of International Concern on January 30th 2020, and a pandemic on March 11th. In Brazil, the first confirmed case was on February 25th 2020, when a man from São Paulo tested positive for the virus. Since then, Brazil has been severely affected. As of June 26th 2020, the country reached more than 1,220,000 confirmed cases and more than 55,000 deaths by COVID-19, according to official data by the Brazilian Ministry of Health.

In addition to the public health crisis, the coronavirus has impacted all aspects of life, politics, education, economy, social, environment and climate. It is also having an unprecedented impact on global supply chains and production.
The only known effective course of action to fight the disease outbreak is to implement highly restrictive social distancing measures on the population, as reported by a number of studies and systematic reviews \cite{Bakker2020, Prem2020, Jefferson2020}. Many countries are implementing such interventions with different degrees of success. 

Given the complexity of the societies, it is hard to predict the implications of such actions in the short and medium terms \cite{ribeiro2020dalmolin}. Therefore, modeling and simulating the COVID-19 epidemic is a relevant and helpful way to understand the spread of the disease and the epidemiological effects of social distancing interventions. For this purpose, many studies in the literature have developed or adapted equation-based models to simulate the COVID-19 epidemic, using the Susceptible-Infected-Recovered (SIR) model or the Susceptible-Exposed-Infected-Recovered (SEIR) model  to characterize the dynamics, see references in Section \ref{sec:referencial}. Nonetheless, agent-based models have also been proposed for this goal, see for instance \cite{ferguson2020report, bossert2020limited} and other studies discussed with more detail in Section \ref{sec:referencial}.

In this paper, we develop an Agent-based Model (ABM) to simulate the dynamics of the COVID-19 epidemic and the epidemiological and economic effects of social distancing interventions. The proposed ABM aims to emulate a closed society living on a shared environment, consisting of agents that represent people, houses, businesses, the government and the healthcare system, each one with specific attributes and behaviors.

A society living over a territory is a complex and dynamic system. Such systems have many interacting variables, present nonlinear behavior and their properties evolve over time. Their behavior is generally stochastic and may depend on the initial conditions. It can be affected by neighbor societies (with different policies and dynamics) and it can show emergence of complex behaviors and patterns. Agent-Based Simulations (ABS) are a good choice to simulate such systems, due to their simplicity of implementation and accurate results when compared with real data \cite{Parunak1998}. 
The main goal of ABS is to simulate the temporal evolution of the system, storing statistics derived from the internal states of the agents in each iteration and the global behaviors that emerge due to the interactions between the agents over the iterations. This approach allows the simulation of systems with intricate nonlinear relationships, complex conditions and restrictions that may be hard to describe mathematically. Since in this paper we are interested in simulating the effects of different social-distancing interventions and other control measures that affect the behaviors of agents and groups of agents, it is much easier to simulate these scenarios with an agent-based model. The epidemiological and economic effects are observed as emerging from the interactions of the agents in the simulation.

The ABM proposed here not only simulates the epidemic dynamics but also models the economy in this society of agents, which can help us estimate the economic impact under different types of interventions. The model (described in Section \ref{sec:model}) allows the design of scenarios that correspond to different types of interventions performed in the society, by changing the simulation environmental variables and measuring their effects. Therefore, the proposed ABM becomes a useful tool to assist politicians and health authorities in planning their actions against the COVID-19 epidemic. The model was implemented in Python version 3.6 programming language and encapsulated in the \texttt{COVID\_ABS package}, whose source code is available at \url{https://bit.ly/COVID19_ABSsystem}. The source code of all the experiments reported herein is available at 
\url{https://bit.ly/covid_abs_experiments}.


The main contributions and findings are listed below:
\begin{itemize}
    \item A new SEIR agent-based model to simulate the COVID-19 epidemic using a society of agents.
    \item Assessment of the economic effects of seven different scenarios with specific social-distancing interventions, via simulation of COVID-ABS: (1) do nothing, (2) lockdown, (3) conditional lockdown, (4) vertical isolation, (5) partial isolation, (6) use of face masks, and (7) use of face masks together with 50\% of adhesion to social isolation. These scenarios and their simulated results are described in Section \ref{sec:results}.
    \item The simulations support the notion that lockdown and conditional lockdown are the best scenarios in terms of controlling the number of infected and deaths, which is primary goal. Economical countermeasures and subsidies are required by the government since this scenario presents the worst economic losses to the industry with potential unemployment, and recession can be observed during the lockdown period. Also, to be effective, these scenarios depend on the ability of the government to enforce the social isolation.
    \item Our simulations present additional evidence that the so-called vertical isolation simply does not work, although it is the policy advocated by some governments like the Brazilian one\footnote{\url{https://agenciabrasil.ebc.com.br/en/politica/noticia/2020-04/bolsonaro-brazil-must-not-be-informed-through-panic}}.
    \item The scenario combining the use of masks and partial isolation of the population could be a good compromise and it is more realistic for implementation in terms of social cooperation. The infection curve is flattened and the economy has smoother effects than the scenarios with lockdown.
\end{itemize}


The rest of the paper is organized as follows: Section \ref{sec:referencial} provides a brief review of related work with focus on the mathematical modeling for epidemics and some recent papers related to the SARS-CoV-2. Section \ref{sec:model} details the proposed agent-based system modeling. Section \ref{sec:exp-methodology} describes the experimental methodology and Section \ref{sec:results} shows the simulations results and some discussions related to the pandemic. Section \ref{sec:conclusion} concludes the paper and gives future directions. 

\section{Related Work}
\label{sec:referencial}

Since WHO announced the Coronavirus Disease 2019, the scientific community has been working hard to investigate SARS-CoV-2 epidemiological dynamics. Some works used the SIR model to characterize the COVID-19 dynamics \cite{anastassopoulou2020data, barlow2020accurate, weissman2020locally,fanelli2020analysis}. However, more precise simulations usually used an approach based on the SEIR model \cite{choi2020estimating, vega2020lockdown, kuniya2020prediction, kim2020school, sugiyanto2020mathematical, manchein2020strong, tang2020effectiveness, tuite2020mathematical, abdo2020comprehensive, maugeri2020estimation, ivorra2020mathematical, liu2020modeling, peirlinck2020outbreak, chatterjee2020healthcare, li2020preliminary, arino2020simple, wang2020phase, wu2020nowcasting}; given that this disease has a known incubation period \cite{lauer2020incubation}. Some authors added new states to refine the model, for instance, super-spreaders \cite{ndairou2020mathematical} or isolated \cite{tang2020effectiveness, vega2020lockdown, li2020preliminary, chatterjee2020healthcare, liu2020modeling, ivorra2020mathematical, tuite2020mathematical, manchein2020strong, kim2020school, choi2020estimating}, hospitalized \cite{tang2020effectiveness, vega2020lockdown, li2020preliminary, ivorra2020mathematical, tuite2020mathematical, choi2020estimating}, and asymptomatic infected \cite{arino2020simple, ivorra2020mathematical, abdo2020comprehensive, tuite2020mathematical, manchein2020strong}.


We note that equation-based models to simulate the epidemic represent the majority among those proposed in the literature. Nonetheless, some papers with agent-based models have also been proposed for it. 
For a discussion about ABM and its advantages over equation-based models, we refer the reader to \cite{Parunak1998, Figueredo2014}.
In the report released by \citet{ferguson2020report}, an individual-based simulation model was used to explore scenarios for COVID-19 in GB and USA and the impact of non-pharmaceutical interventions on the healthcare demand.
\citet{bossert2020limited} developed an agent-based model combining socio-economic and traffic data to analyze COVID-19 spreading in a South Africa city under social isolation scenarios. The prediction suggests that lockdown strategy is useful to mitigate the disease. Another study using an ABM also analyzed several scenarios and highlighted that with 90\% of the population in isolation, it is possible to control the disease within 13 weeks when joined with effective case isolation and international travel restrictions, considering the Australian context \cite{chang2020modelling}. An appealing characteristic of agent-based modeling is the easiness to simulate different scenarios. For instance, the scenario that considers universal use of masks integrated with social distance is the recommended one to control the pandemic according to \citet{braun2020phase} and \citet{kai2020universal}. Given the flexibility of the agent-based approach, previous works have employed this method to simulate specific topics in the COVID-19 context, such as testing policies \cite{gopalan2020reliable}, strategies for reopening public buildings \cite{d2020restart}, hypothetical effective treatments \cite{hoertel2020facing}, and a spatio-temporal strategy for vaccination \cite{grauer2020strategic}.

Few works in the literature used agent-based models to simulate the economic impacts of the COVID-19. For instance,  \citet{inoue2020propagation} quantified that a possible one month lockdown in Tokyo would lead to a total production loss of 5.3\% in Japanese annual gross domestic product (GDP). \citet{dignum2020analysing} proposed a tool to analyze the health, social, and economic impacts of the pandemic when the government implements a number of interventions, such as  closing schools, requiring that employees work at home, and providing subsidy for the population.

In this work, we use a SEIR agent-based model to simulate the health and economic impacts of the COVID-19 epidemic. We perform analysis to seven possible scenarios: (1) do nothing, (2) lockdown, (3) conditional lockdown, (4) vertical isolation, (5) partial isolation, (6) use of face masks, and (7) use of face masks together with 50\% of social isolation. We use data from Brazil for all scenarios considered but the proposed agent-based model is fully parameterized and can be easily transferred to other contexts given that corresponding data is provided. 
\section{COVID-ABS: Proposed Agent-based System Modeling}
\label{sec:model}

The proposed agent-based approach aims to emulate a closed society living on a shared finite environment, composed of humans, which are organized in families, business and government, which interact with each other. This characterization is trying to cover the main elements of the society. The agents, their attributes and possible actions are described in Table \ref{tab:agent_type}.

\begin{table}[h!]
\small
\centering
\begin{tabular}{p{2.2cm} p{11cm} }
\hline
\multicolumn{2}{c}{\textbf{A1: Person}} \\ \hline
{Description} & $A1$ is the main type of agent. Its dynamic position varies according to the environment and may be associated with $A2$, or not (homeless) and $A3$, or not (unemployed). \\ 
{Attributes} & Position (dynamic),  Age,  House ($A2$),  Employer  ($A3$),  Epidemiological  status,  Infection  status,   Wealth,   Income   and   Social Stratum \\ 
{Actions} & Walk freely (daily),  Go home (daily),  Go to work  (daily),  Personal contact (hourly),   Business contact (hourly), Go to the hospital  \\ \hline
\multicolumn{2}{c}{\textbf{A2: Houses}} \\ \hline
{Description} & $A2$ represent the families. They share a house and financial bills. \\ 
{Attributes} & Position (static), Social stratum, Housemates  (group  of  $A1$), Wealth, Incomes and Expenses \\ 
{Actions} & Homemate check-in (daily),   Accounting (monthly) \\ \hline
\multicolumn{2}{c}{\textbf{A3: Business}} \\  \hline
{Description} & $A3$ are the economical agents, e.g. industries, shops or markets. It interacts with $A1$ by paying a salary or selling a product. \\ 
{Attributes} & Position (static), Social stratum, Employees (group of $A1$), Wealth, Incomes and Expenses \\ 
{Actions} & Accounting (monthly), Business contact (hourly) \\ \hline
\multicolumn{2}{c}{\textbf{A4: Government}} \\ \hline
{Description} & $A4$ is a singleton agent that receives taxes from $A2$ and $A3$, provide funds to $A5$ and insurance for homeless and unemployed $A1$. \\ 
{Attributes} & Position (static), Wealth \\ 
{Actions} & Accounting (monthly) \\ \hline
\multicolumn{2}{c}{\textbf{A5: Healthcare System}} \\  \hline
{Description} & $A5$ is also a singleton, which represents the health system that ideally should be able to serve the entire population. \\ 
{Attributes} & Position (static), Wealth  \\ \hline
\end{tabular}
\caption{Types of agents and their attributes and actions.}
\label{tab:agent_type}
\end{table}

The model is an iterative procedure, with $T$ representing the number of iterations. The model takes an input parameter set $P$, listed in Table \ref{tab:model_variables}, and produces a response $Y_t$ (observable variables), related to epidemiological or economic effects of the pandemic. Its internal state $\Theta_t$ ($t=1 \hdots T$) consists of the union of the internal states of the agents $\theta^i_t$, where $i=1\ldots n$ and $n$ is the number of agents, such that $\Theta_t = \bigcup_{i=1}^n \theta^i_t$. 

\begin{table*}[h!]
\footnotesize
\begin{tabular}{p{0.3\textwidth} p{0.11\textwidth} p{0.11\textwidth} p{0.37\textwidth}}
\hline
\textbf{Variable} & \textbf{Domain/ Unit} & \textbf{Current value} & \textbf{References and Observations} \\ \hline
\multicolumn{4}{c}{\textbf{Social and Demographic}} \\ \hline
$\alpha_1$ - Height & ${\mathbb{N}^+}$ & $500$ & Defined empirically. Each unit corresponds to 7 meters. \\
$\alpha_2$ - Width & ${\mathbb{N}^+}$ & $500$ & Defined empirically. Each unit corresponds to 7 meters. \\
$\alpha_3$ - Population size & ${\mathbb{N}^+}$/people & $300$ & Defined empirically. \\
$\alpha_4$ - Age & $[0, 100]$ & $\beta (2, 4)$ & \cite{IBGEfaixaetaria} \\
$\alpha_5$ - Average family size & ${\mathbb{N}^+}$/people & $3$ & \cite{IBGE_MoradoresDomicilios} \\
$\alpha_6$ - Mobility & ${\mathbb{N}^+}$ & $10$ & Defined empirically.  Each unit corresponds to 7 meters. \\
$\alpha_7$ - Homeless rate & $[0, 1]$ & $0.0005$ & \cite{IPEA_Homeless} \\ \hline
\multicolumn{4}{c}{\textbf{Epidemiological}} \\ \hline
$\beta_1$ - Contagion distance & ${\mathbb{R}^+}$ & $1$ & \cite{ImperialCollege_Contagiondistance} \\
$\beta_2$ - Contagion probability & $[0, 1]$ & $0.9$ & \cite{ImperialCollege_Contagiondistance} \\
$\beta_3$ - Incubation time & ${\mathbb{N}^+}$/days & $5 - 6$ & \cite{lima2020_tempoincubacao,li2020_tempoincubacao} \\
$\beta_4$ - Transmission time & ${\mathbb{N}^+}$/days & $8 - 10$ & \cite{lauer2020_incubation_transmissao} \\
$\beta_5$ - Recovering time & ${\mathbb{N}^+}$/days & $20$ & \cite{Portal_TheConversation} \\
$\beta_6$ - Hospitalization rate per age & $[0, 1]$ & Table \ref{tab:rates_medicalconditions} &  \cite{ImperialCollege_Contagiondistance}\\
$\beta_7$ - Severe cases rate per age & $[0, 1]$ & Table \ref{tab:rates_medicalconditions} &  \cite{ImperialCollege_Contagiondistance}\\
$\beta_8$ - Death rate per age & $[0, 1]$ & Table \ref{tab:rates_medicalconditions} &  \cite{ImperialCollege_Contagiondistance}\\
$\beta_9$ - \% initial infected & $[0, 1]$ & $0.01$ & Defined by the authors. \\
$\beta_{10}$ - \% initial immune & $[0, 1]$ & $0.01$ & Defined by the authors.  \\
$\beta_{11}$ - Critical limit of the Health System & $[0, 1]$ & $0.05$ & The proportion of ICU beds to the population \\ \hline
\multicolumn{4}{c}{\textbf{Economical}} \\ \hline
$\gamma_1$ - Income distribution &  & Table \ref{tab:wealth_distribution} &  \cite{Portal_IndexMundi,WorldBank}\\
$\gamma_2$ - Proportion of businesses & ${\mathbb{R}^+}$ & 0,01875  &  Considering the number of businesses per 100k inhabitants \cite{IBGE_demografiaEmpresas} \\
$\gamma_3$ - Total GDP & ${\mathbb{R}^+}$/R\$ & $1.000.000,00$ & Defined by the authors. \\
$\gamma_4$ - Public GPD rate & $[0, 1]$ & $0.01$ & Defined by the authors. \\
$\gamma_5$ - Business GPD rate & $[0, 1]$ & $0.05$ & Defined by the authors. \\
$\gamma_6$ - Personal GPD & $[0, 1]$ & $0.04$ & $\gamma_6 = 1 - \% A4 - \%A3$ \\
$\gamma_7$ - Minimum income & ${\mathbb{R}^+}$/R\$ & $900,00$ &  \\
$\gamma_8$ - Minimum expenses & ${\mathbb{R}^+}$/R\$ & $600,00$ &  \\
$\gamma_9$ - Unemployment rate & $[0, 1]$ & $0.12$ & \cite{EXAME_desemprego} \\
$\gamma_{10}$ - Proportion of informal businesses & $[0, 1]$ & $0.40$ & Informal economy \cite{EXAME_desemprego,Sebrae_cnpjInformal} \\
$\gamma_{11}$ - EAP age group 
&  & & $16<\text{EAP}<65$  \\ \hline
\end{tabular}
\caption{Definitions of the parameters of the proposed ABS model}
\label{tab:model_variables}
\end{table*}

The model is described in Algorithm 1. The initialization of internal states in line \ref{alg:linha1}, discussed in subsection \ref{subsec:initial}, creates the agents. The simulation dynamics starts in line \ref{alg:linha2}, discussed in subsection \ref{subsec:dynamics}, and depends on the type of the agent, the parameter $P$ and the current iteration $t$ (discrete time). As mentioned before, each type of agent has its own set of actions in different time frames (hourly, daily, weekly or monthly).

\setcounter{algorithm}{0}
\begin{algorithm}
\begin{algorithmic}
\Require{$P$ the parameter set, $T$ the number of iterations}
\State {$\Theta_0 \gets$ \texttt{initialize}($P$)} \label{alg:linha1}
\For{$t \gets 1 \textrm{ to } T$}   \label{alg:linha2}
    \ForAll{agent $a_i \in \Theta_t$}
        \State {$\theta^i_t \gets a_i$.\texttt{execute\_actions}($t, P, \Theta_t$)}
        \If {type of $a_i = A1$ }
        \ForAll{agent $a_j \in \Theta_t ~|~ i \neq j$}
            \If {distance($a_i, a_j$) $\leq \delta$}
                \State {$a_i$.\texttt{contact}($a_j$)}
            \EndIf
        \EndFor
        \EndIf
    \EndFor
    \State {$Y_t \gets$ \texttt{summarize}($\Theta_t$)}
        \State {$\Theta_{t+1} \gets \bigcup_{i=1}^n \theta^i_t$}
\EndFor
\end{algorithmic}
\label{alg:covid_abs1}
\caption{General procedure of the proposed agent-based approach}
\end{algorithm}

At each iteration, it checks if there was contact between any pair of agents. A contact happens when the distance between any two agents is less than or equal to a threshold $\delta$ defined in $P$. The contact can be epidemiological (if the agents are of type A1) or economical (A1 and A3). The computation of the distance between each pair of agents per iteration makes the asymptotic complexity of the method equal to $O(n^2)$, where $n$ is the number of agents.


\begin{table}[htb]
    \centering
    \begin{tabular}{c p{11cm}}  \hline
         \textbf{Variable} & \textbf{Description}  \\ \hline
         \multicolumn{2}{c}{\textbf{Epidemiological}} \\ \hline
         $S_t$ & Percentage of Susceptible agents in population \\ \hline
         $I_t$ & Percentage of Infected agents in population \\ \hline
         $I^A_t$ & Percentage of Infected Asymptomatic agents in population \\ \hline
         $I^H_t$ & Percentage of Infected Hospitalized agents in population \\ \hline
         $I^S_t$ & Percentage of Infected Severe agents in population \\ \hline
         $R_t$ & Percentage of Recovered and Immune agents in population \\ \hline
         $D_t$ & Percentage of Dead agents in population \\ \hline
         \multicolumn{2}{c}{\textbf{Economical}} \\ \hline
         $W^{A1}_{S,t}$ & Percentage of Gross Domestic Product owned by the people (A1 agents) at time $t$ under scenario $S$ \\ \hline
         $W^{A3}_{S,t}$ & Percentage of Gross Domestic Product owned by businesses (A3 agents) at time $t$ under scenario $S$ \\ \hline
         $W^{A4}_{S,t}$ & Percentage of Gross Domestic Product owned by government (A4 agent) at time $t$ under scenario $S$ \\ \hline
    \end{tabular}
    \caption{Response Variables}
    \label{tab:response_variables}
\end{table}

\subsection{Parameter estimation}
\label{subsec:par_estim}

Some parameters in Table \ref{tab:model_variables} were empirically estimated, in a way that the epidemiological response variables present in Table \ref{tab:response_variables} correspond to those produced by a SEIR model. For that purpose the Epidemic Calculator\footnote{https://gabgoh.github.io/COVID/} was employed -- this is an open-source SEIR implementation and visual tool for epidemic simulations. The initial percentage of infected ($\beta_9$) and immune ($\beta_{10}$) agents were chosen in order to represent the complete epidemic dynamics.

The population size parameter, $\alpha_3$, is particularly concerning because it affects the execution time of the simulation. On the other hand, the Population density, defined as $\alpha_3/(\alpha_1  \alpha_2)$, follows the population density of the area under study which is 24 people per km$^2$. The mobility parameter $\alpha_6$ was empirically estimated as the average range that a person walks randomly in his free time.

The Total GDP, parameter $\gamma_3$, and the percentage rates by kind of agent ($\gamma_4, \gamma_5$ and $\gamma_6$) are abstractions of closed local economy. The minimum income $\gamma_7$ represents the minimum net salary, the nominal income after taxes, and the minimum expense $\gamma_8$ represents the approximate market value of a basic needs grocery pack.

\subsection{Initialization}
\label{subsec:initial}

The simulation is performed in a squared bi-dimensional environment shared by all types of agents. $Ai$ agents, $i \in \{2,3,4,5\}$,  are randomly initialized inside this environment given by Equation \eqref{eq:agent_initial_position}.

\begin{equation}
\label{eq:agent_initial_position}
\centering Ai_{pos} = \left\{
\begin{array}{ll} 
x \sim \mathcal{U}(0, \alpha_1) \\ 
y \sim \mathcal{U}(0, \alpha_2)
\end{array}\right.
\end{equation}
where $\mathcal{U}(a,b)$ is a sample from a uniform distribution in the interval $[a,b)$.

Agents $A1$ are initialized in their $A2$ location, following Equation \eqref{eq:person_initial_position}, where $\sigma_k$ is the variability of the position inside the house. For homeless agents, Equation \eqref{eq:agent_initial_position} is used.  

\begin{equation}
    A1_{pos} = A2_{pos} + \mathcal{N}(0, \sigma_k)
    \label{eq:person_initial_position}
\end{equation}
where $\mathcal{N}(\mu,\sigma)$ is a sample from a normal distribution with mean $\mu$ and standard deviation $\sigma$.

The number of $A1$ agents is controlled by the variable population size, that is $|A1|= \alpha_3$. The number of houses ($A2$ agents) is calculated using Equation \eqref{eq:houses_amount} considering the average family size $\alpha_5$:

\begin{equation}
    |A2| = \left \lceil \frac{\alpha_3}{\alpha_5} \right\rceil 
    \label{eq:houses_amount}
\end{equation}

The number of $A3$ agents is calculated according to Equation \eqref{eq:total_business}, considering the population size $\alpha_3$, the proportion of formal and informal businesses, $\gamma_2$ and $\gamma_{10}$, respectively.

\begin{equation}
    |A3| = \lceil \alpha_3  \gamma_2 + \alpha_3  \gamma_{10} \rceil 
    \label{eq:total_business}
\end{equation}

When a person, $A1$ type, is created, it is assigned to a randomly chosen house, type $A2$, or it is considered homeless according to the Homeless Rate, $\alpha_7$. Parameter $\gamma_9$ defines the probability of an $A1$ to be unemployed. If a person is employed and belongs to Economical Active Population (EAP) (controlled by $\gamma_{11}$) an employer is randomly chosen among the available $A3$s. A single instance of $A4$ and $A5$ agents are created.

The age distribution of $A1$ agents is given by $\alpha_4$ parameter, such as $A1_{age} \sim\beta(2,5)$ as explained in \cite{HowdenMeyer2010}, where $\beta(a,b)$ is the beta distribution with shape parameters $a,b$. 

The social stratum of $A1, A2$ and $A3$ is represented by the income distribution $\gamma_1$, listed in the Table \ref{tab:wealth_distribution}, meaning the slice of the wealth represented by the GDP parameter $\gamma_3$.  The social stratum of agents is sampled such that $Ai_{stratum} \sim \mathcal{U}(1,5)$, for $i = \{1,2,3\}$. The total wealth of the simulation, represented by $\gamma_3$, is shared among agents, according to public, business and personal percentages defined by $\gamma_4$, $\gamma_5$ and $\gamma_6$.

\begin{table}[h!]
\centering
\begin{tabular}{clrr} \hline
Quintile & Social Stratum & \begin{tabular}[c]{@{}c@{}}\% of GDP \\ Share\end{tabular} & \begin{tabular}[c]{@{}c@{}}Cummulative\\ \% of GDP Share\end{tabular} \\ \hline
Q1 & Most Poor & 3.62 & 3.62 \\
Q2 & Poor & 7.88 & 11.50 \\
Q3 & Working Class & 12.62 & 24.17 \\
Q4 & Rich & 19.71 & 43.88 \\
Q5 & Most Rich & 56.12 & 100.00 \\ \hline
\end{tabular}
\caption{Income distribution ($\gamma_1$).  Adapted from World Bank \cite{WorldBank} }
\label{tab:wealth_distribution}
\end{table}

After the creation of all agents the simulation model starts its iteration loop, which represents the time dynamics, explained in the next section.

\subsection{Simulation Dynamics}
\label{subsec:dynamics}

\begin{figure}[h!]
    \centering
    \includegraphics[width=\textwidth, height=6cm]{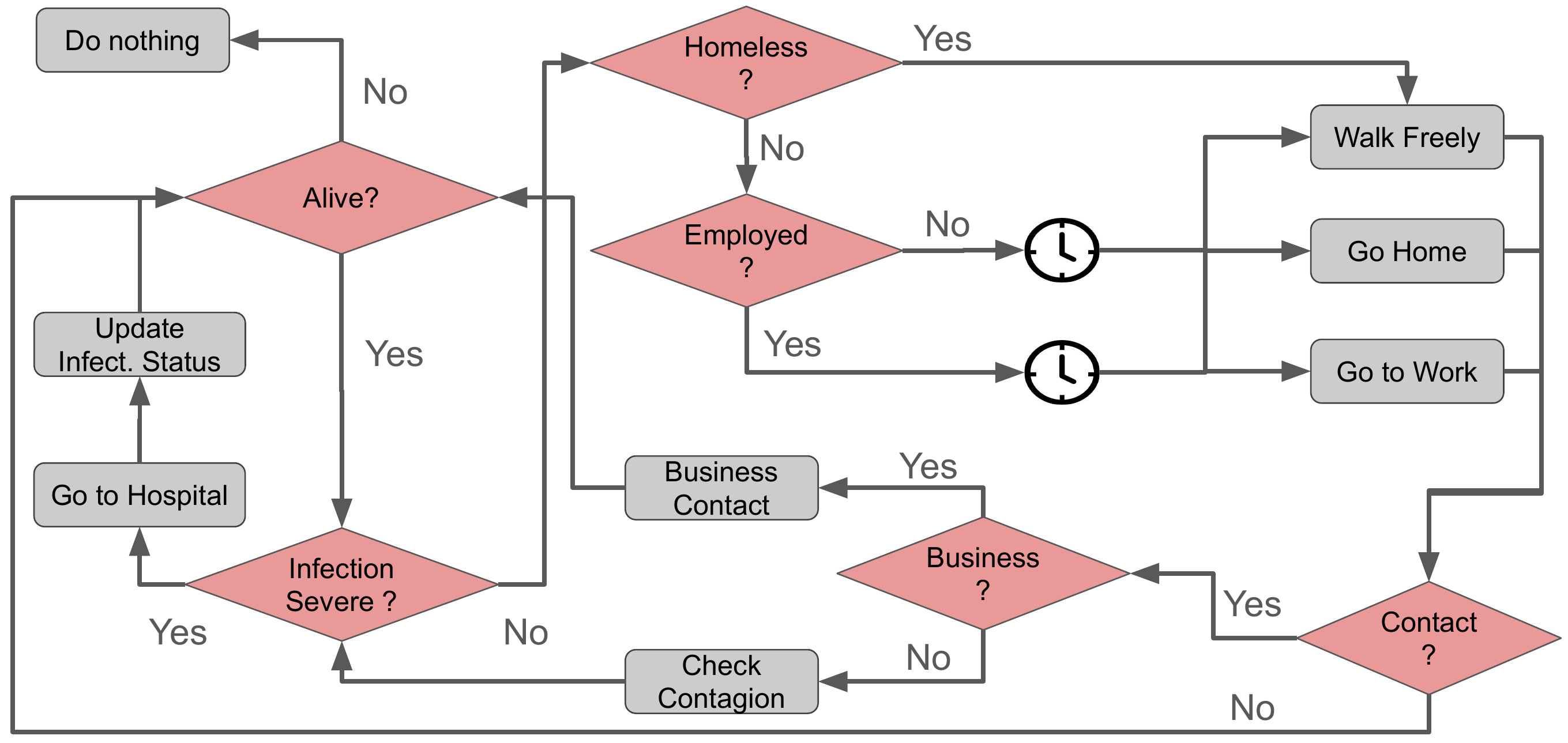}
    \caption{A1 agent activity cycle }
    \label{fig:people_fluxogram}
\end{figure}

Each iteration represents one hour when the agents are invoked to perform actions that depend on their type and behaviors, as shown in Figure \ref{fig:people_fluxogram} for $A1$ agents, and more detailed in subsection \ref{subsec:mobility}. During its movement, an $A1$ agent may get in the proximity with other $A1$, $A2$ or $A3$ agents. Subsection \ref{subsec:spreading} presents the possibility of contagion that can happen through contact between two $A1$ agents. Finally, subsection \ref{subsec:economicalimpacts} presents the economic relationships between agents, caused by contact of $A1$ and $A3$ agents (commercial transactions), payment of taxes for the government ($A4$ agent), labor relationships between $A3$ and $A1$ agents and house expenses between $A2$ and $A1$ agents.

\subsubsection{$A1$ Mobility Patterns}
\label{subsec:mobility}

The distribution of $A1$ agents' work, rest and leisure hours is shown in Table \ref{tab:mobility} and it is based on the Universal Declaration of Human Rights \cite{ohchr}. Basically, it is the standard deviation of a Gaussian distribution with average $\mu = 0$, representing the variability of the movement amplitude of $A1$ in its free time or, in other words, how far the agent can walk from its actual position.

\begin{table}[!htb]
\small
\centering
\begin{tabular}{cccl}
\hline
\multicolumn{1}{c}{\textbf{Start Time}} & \multicolumn{1}{c}{\textbf{End Time}} & \multicolumn{1}{c}{\textbf{Activity}} &
\multicolumn{1}{c}{\textbf{Action}} \\ \hline 
0 & 8 & Rest & \begin{tabular}{l}If $A1$ is not homeless: \\ \hspace{1cm}Go home (Equation \eqref{eq:move_to_home})\\ Otherwise: \\  \hspace{1cm}Walk freely (Equation \eqref{eq:move_freely})\end{tabular} \\  \hline
8 & 12 & Job & \begin{tabular}{l}If $A1$ is not unemployed: \\ \hspace{1cm}Go to work (Equation \eqref{eq:move_to_work})\\ Otherwise: \\  \hspace{1cm}Walk freely (Equation \eqref{eq:move_freely})\end{tabular} \\  \hline
12 & 14 & Lunch & Walk Freely (Equation \eqref{eq:move_freely}) \\ \hline
14 & 18 & Job & \begin{tabular}{l}If $A1$ is not unemployed: \\ \hspace{1cm}Go to work (Equation \eqref{eq:move_to_work})\\ Otherwise: \\  \hspace{1cm}Walk freely (Equation \eqref{eq:move_freely})\end{tabular} \\  \hline
18 & 0 & Recreation & Walk freely (Equation \eqref{eq:move_freely}) \\ \hline
\end{tabular}
\caption{A1 agent movement routines considering a full day and different activities}
\label{tab:mobility}
\end{table}

The actions ``Go home'', ``Go to work'' and ``Walk freely'' occur according to the Equations \eqref{eq:move_to_home}, \eqref{eq:move_to_work} and  \eqref{eq:move_freely}. Besides these ordinary actions, all the agents that are infected and have infection severity equal to hospitalization or severe execute the ``Go to hospital'' action, according to the Equation  \eqref{eq:move_to_hospital}. All the dead agents  have their positions set to zero.

\begin{flalign}
    &A1_{pos} = A2_{pos} + \mathcal{N}(0,\sigma_k) \label{eq:move_to_home}\\
    &A1_{pos} = A3_{pos} + \mathcal{N}(0,\sigma_k) \label{eq:move_to_work}\\
    &A1_{pos} = A1_{pos} + \mathcal{N}(0, \alpha_6) \label{eq:move_freely}\\
    &A1_{pos} = A5_{pos} + \mathcal{N}(0,\sigma_k) \label{eq:move_to_hospital}
\end{flalign}
where $\sigma_k = 0.01$ is the random noise variance for ``Go to...'' actions, and the mobility parameter $\alpha_6$ is the random noise variance for ``Walk freely'' action, representing the amplitude of movement the A1 agents have in their free time.

\subsubsection{Contagion Spreading}
\label{subsec:spreading}

COVID-19 is a highly contagious disease. According to the Report 3 of the Imperial College London ``on average, each case infected 2.6 (uncertainty range: 1.5–3.5) other people up to 18th January 2020'' \cite{imai2020report}. Following the SEIR model, in each simulation, there is an initial percentage of infected and immune people ($\beta_9$ and $\beta_{10}$, respectively), and the remaining population consists of susceptible individuals. There is also a Dead status, since part of the population dies due to the disease and its complications \cite{giordano2020Italy}. 

The possibility of contagion happens by the interaction of the agents by proximity or contact. Hence, the higher the mobility of a person, the greater  the probability that he/she approaches an infected person and gets infected. Each simulation considers a contagion distance threshold $\beta_1$, which is the minimal distance that two agents have to be to occur the viral transmission, and a probability of contagion $\beta_2$ in case of contact.

\begin{figure}[h!]
    \centering
    \includegraphics[width=\textwidth,height=6cm]{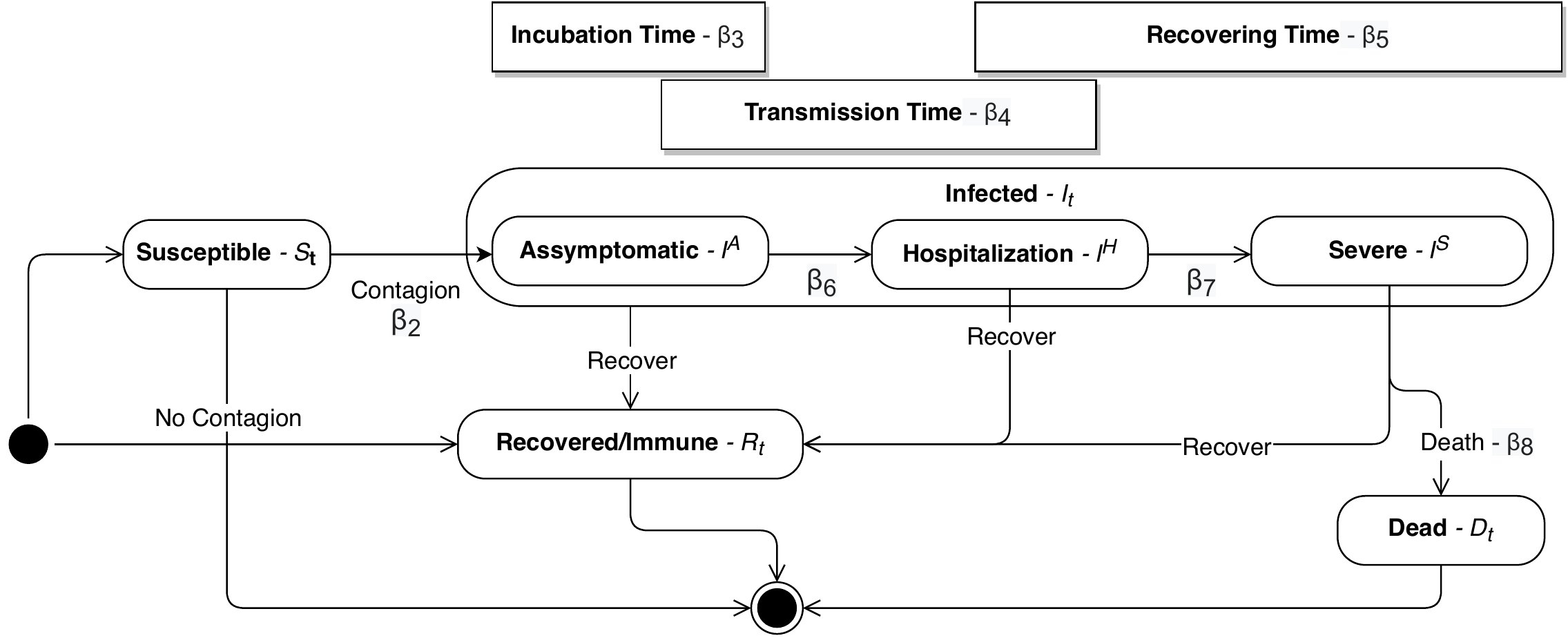}
    \caption{Epidemiological and infection state diagram for A1 agents based in SEIR model, with the corresponding population response variables and parameters of their transition probabilities }
    \label{fig:state_transition}
\end{figure}

The model of the medical condition evolution of the infected agents follows  \cite{verity2020estimates,dorigatti2020report}. Once an agent is infected, it can be in one of these sub-states: a) asymptomatic, which includes mild symptoms without hospitalization, b) hospitalization and c) severe, used in cases of hospitalization in intensive care unit (ICU). These states and their transitions are illustrated in Figure \ref{fig:state_transition}.

The evolution of the medical condition is stochastic and follows the probabilities summarized in Table \ref{tab:rates_medicalconditions}, represented by the parameters $\beta_6$, $\beta_7$ and $\beta_8$, respectively. The hospitalization cases require medical infrastructure, which is limited. It varies from country to country, but is always less than the total population. In each simulation, a critical limit $\beta_{11}$ is considered, it represents the percentage of the population that the healthcare system is capable to handle simultaneously. As a consequence, if the number of hospitalizations and severe cases increase above this limit, there are no beds in hospitals to manage the demand.

\begin{table}[h!]
\centering
\resizebox{\textwidth}{!}{
\begin{tabular}{c r r r}
\hline
\textbf{\begin{tabular}[c]{@{}c@{}}Age-group\\ (years)\end{tabular}} & \textbf{\begin{tabular}[c]{@{}c@{}}$\beta_6$ - \% symptomatic cases\\ requiring hospitalization\end{tabular}} & \textbf{\begin{tabular}[c]{@{}c@{}}$\beta_7$ - \% hospitalised cases\\ requiring critical care\end{tabular}} & \textbf{\begin{tabular}[c]{@{}c@{}}$\beta_8$ - Infection Fatality\\ Ratio\end{tabular}} \\ \hline
{0 - 9} & 0.100 & 5.000 & 0.002 \\ 
{10 - 19} & 0.300 & 5.000 & 0.006 \\ 
{20 - 29} & 1.200 & 5.000 & 0.030 \\ 
{30 - 39} & 3.200 & 5.000 & 0.080 \\ 
{40 - 49} & 4.900 & 6.300 & 0.150 \\ 
{50 - 59} & 10.200 & 12.200 & 0.600 \\ 
{60 - 69} & 16.600 & 27.400 & 2.200 \\ 
{70 - 79} & 24.300 & 43.200 & 5.100 \\ 
{80+} & 27.300 & 70.900 & 9.300 \\ \hline
\end{tabular}}
\caption{Rates of medical conditions considering hospitalized ($\beta_6$) and severe ($\beta_7$) and death ($\beta_8$) cases grouped by age. Adapted from \citet{ferguson2020report}}
\label{tab:rates_medicalconditions}
\end{table}

\subsubsection{Economic Transactions}
\label{subsec:economicalimpacts}

The secondary goal of this study is to simulate the impact caused in the economy by the different types of mobility restrictions  \cite{singh2020age,meidan2020alternating,warren2020mobility,engle2020staying} imposed by the authorities.

\begin{figure}[!ht]
    \centering
    \includegraphics[width=9cm]{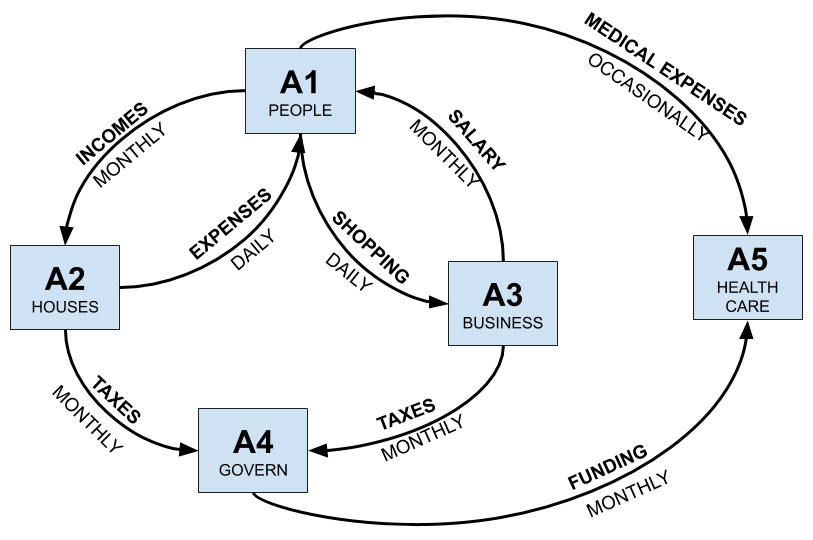}
    \caption{Economic relationships between agents.}
    \label{fig:agent_ecom_relation}
\end{figure}

Figure \ref{fig:agent_ecom_relation} shows the transactions by which agents exchange wealth in the simulation. The economic dynamics follows seasonal routines that also depend on the type of the agent. 

The ``business contact'' action happens hourly, when an A1 agent in its free time gets in contact with an A3, and occurs the transference of wealth from A1 to A3. These economic transactions are the most sensitive to the A1 agents mobility (the more the agents move, the more they spend) and affects the A3 agent income. In pandemic times, that can happen in almost all scenarios, since the population tends to leave their houses just to buy essential items or to solve a problem which could not have been solved over the Internet. The values exchanged in ``business contact'' depend on the social stratum of the A1 agents, and the higher the quintile the higher the spending following the wealth distribution $\gamma_1$. In each day, the wealth of A2 and A3 agents is decreased by its minimal fixed expenses, proportional to the sum of the expenses of housemates and employees, respectively.

The ``accounting'' actions happen monthly for A2, A3 and A4 agents. Accounting is the payment of taxes from A2 and A3 agents to A4, and it represents the major income of A4. During accounting, A3 agents also pay salaries to their A1 employees determined in the initialization by the social stratum. Finally, A2 agents transfer money to a random A3 agent, representing supplier payments.

The accounting of the government agent, A4, transfers funds to A5 agent, equivalent to its fixed expenses and the daily expenses of the hospitalized agents. Eventually, the A4 agent pays aids for unemployed and homeless A1 agents.

Considering the periodicity of the economic transactions, it is necessary to execute at least one complete cycle (720 iterations) in order to execute all economic transactions at least once. 

\subsection{Discussion}

The proposed model tries to apprehend the complexity of the social, epidemiological and economical relationships, but without being simplistic. Then, the complexity of the model reflects the complexity of the system being modelled although on a smaller scale.

Previous versions of the model did not consider social constraints and routines and, despite the good performance of the epidemiological response variables, it could not accurately represent the economic dynamics. The introduction of social constructs, as families and businesses, and periodical routines, such as working hours, free time and bed time, brought more feasibility to the simulation and improved the performance of the economic response variables when compared with the real world values. 

These social constructs and periodic routines are hard-coded, although they can be adapted by users. Major flexibility is provided by the parameter set that can be adjusted to represent from a single community to a complete country. 

In the next section, the experimental methodology is discussed, as well as the performance metrics and their evaluation.

\section{Experimental Methodology}
\label{sec:exp-methodology}

To evaluate the proposed approach, seven different scenarios that reflect adopted and/or hypothetical social distancing interventions have been formulated. The proposed ABS model was implemented in Python version 3.6 programming language and encapsulated in the \texttt{COVID\_ABS package}, whose source code is available at \url{https://bit.ly/covid_abs_experiments} and \url{https://bit.ly/COVID19_ABSsystem}.


Each scenario simulates the impact of a given social distancing policy, considering the values of the parameters in Table \ref{tab:model_variables}, on the response variables, summarized in Table \ref{tab:response_variables}. 
For each scenario 35 executions were performed, each one with $T = 1,440$ iterations. Since each iteration corresponds to one hour, each execution covers exactly $2$ months and one complete accounting cycle for houses, government and business, with one salary and tax payment, which occur in the 30th day of the month. The monthly ``accounting'' event is important for $A2$, $A3$ and $A4$ agents due to its severe cash impact and wealth transfers among agents.


The main goal of social interventions is to minimize the death curve $D_t$. It is directly related to flattening the infection curve $I_t$, in order to keep the hospitalization $I^H_t$ and severe $I^S_t$ cases below the critical limit of the healthcare system $\beta_{11}$. Flattening the $I_t$ curve means minimizing the infection peak $I_{P}$, defined in Equation \eqref{eq:infection_peak}, and extending the time $T_{IP}$ spent to reach this peak, defined in Equation \eqref{eq:days_to_peak}.

\begin{align}
    I_{P} = \max \{\; I_t\; |\; t=1\ldots T\; \}
    \label{eq:infection_peak} \\
    T_{IP} = \min\{\; t\; |\; I_t = I_{P}\; \} 
    \label{eq:days_to_peak}
\end{align}

To compare the scenarios, the response variables $D_t$, $I_t$ will be considered, condensed in the metrics $I_P$ and $T_{IP}$.

The economical analysis aims to assess the evolution of wealth, represented by the $W^*_{S,t}$ response variables. To allow the economic comparison among scenarios with respect to the same reference, a baseline scenario, $B$, without a pandemic was designed. It is meant to isolate the economic dynamic and can be used to assess the impacts of the different interventions in the economy.

For comparison among scenarios, the increase in wealth, $\Delta W^i_S$, for the group of agents, $i \in \{A1, A3, A4 \}$, in scenario $S$ is computed as follows:

\begin{equation}
\Delta W^i_S  =  \frac{W^i_{S,T} - W^i_{B,T}}{W^i_{B,T}} 
    \label{eq:wealth_result}
\end{equation}

\noindent where $W^i_{B,T}$ is the wealth of the group of agents, $i$, at the final simulation time step, $T$, of $B$. 

In the following section, the scenarios are defined, their simulation results are presented and compared, and the main findings discussed.
\section{Results and Analysis}
\label{sec:results}

This section shows the results for seven scenarios, chosen to represent the major interventions adopted or defended by governments. In this study, the parameters are based on data from Brazil. Nevertheless, in order to encourage further studies and/or applications, it is possible to easily transfer the model to other societies by changing the social, demographic and economical parameters listed in Table \ref{tab:model_variables} and creating a multitude of other scenarios adapted to the new regions. 

The meaning of each scenario, its parameters and dynamics of the response variables are discussed below. 

\subsection{Baseline ($B$): No Coronavirus Pandemic}
\label{subsec:result_scenario_0}


This scenario simulates the economic behavior without a pandemic. It is used as baseline for comparison with all the other scenarios. To generate this scenario, set parameters to $\beta_9 \leftarrow 0$ and $\beta_{10} \leftarrow 1$.

Despite the economic result of this simulation, it is artificial data and consequently it does not represent the reality of any country. Nonetheless, we argue that it is based on projections before the pandemic outbreak. See, for instance, the references regarding each parameter listed in Table \ref{tab:model_variables}. 

\begin{figure}[h!]
    \centering
    \includegraphics[width=\textwidth, height=6cm]{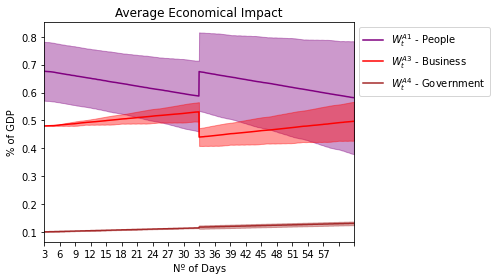}
    \caption{Daily averaged response variables for $B$.}
    \label{fig:scenario0}
\end{figure}

The evolution of GDP is illustrated in Figure \ref{fig:scenario0}. The GDP indicates a recession chart where the population ($A1$) and government ($A4$) is losing wealth and the businesses ($A3$) are floating at the equilibrium point (when the incomes and expenses are equal). Initially the $A3$ are profiting but, in the accounting day, the profits are settled by the labor and tax expenses. The baseline scenario is consistent with the economic predictions of stagnation in Brazil.

\subsection{Scenario 1: Do Nothing}
\label{subsec:result_scenario_1}

\begin{figure}[h!]
    \centering
    \includegraphics[width=1.1\textwidth, height=4.5cm]{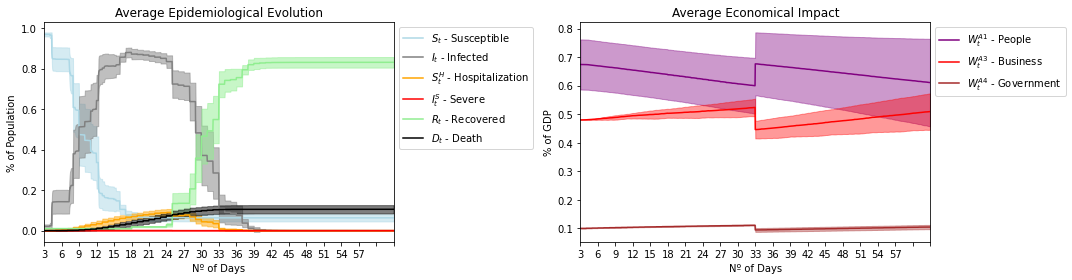}
    \caption{Daily averaged response variables for scenario ``Do Nothing''} 
    \label{fig:scenario1}
\end{figure}

This scenario represents what could happen if politicians decided not to take any actions to avoid the increase of the number of people infected by the SARS-CoV-2 virus. Usually, this decision only targets the economic point of view. Figure \ref{fig:scenario1} shows the epidemiological and economical average curves of this scenario and their variances. It can be seen that the economic curves look closer to the ones of the baseline, confirming the economic motivation of keeping the environment without interventions.

However, when the contagion curve $I_t$ is considered, it is possible to note how the Healthcare System critical limit $\beta_{11}$ was trespassed, pushing the death curve $D_t$ up. The high number of lost lives makes this the most  catastrophic scenario, despite its economic resemblance with $B$. 

\subsection{Scenario 2: Lockdown}
\label{subsec:result_scenario_2}

This scenario represents the complete social isolation, following the WHO recommendations, during a well defined date range. In this scenario, all $A1$ agents are kept in their houses, and the ``walk freely'' and ``go to work'' routines are suppressed. Also $\alpha_6 \gets 1$, reducing the mobility amplitude of all $A1$ even the homeless, as discussed in Section \ref{subsec:mobility}. The lockdown is unconditional, meaning that from $t=0$ to $T$, all the restrictions are applied.

This scenario is highly conservative in healthcare terms, and the main goal is to save as many lives as possible by minimizing viral spreading. In the impossibility of effective testing, the entire population stays in lockdown for a predefined period of time. Broadly speaking, the infected agents only have contact with their housemates and the $I_t$ (and especially $I^S_t$) stays below the healthcare critical limit $\beta_{11}$, and the deaths $D_t \gets 0$, meaning that the healthcare system could handle effectively all cases, using its available resources\footnote{Considering the given population size in the simulation.}. 

\begin{figure}[h!]
    \centering
    \includegraphics[width=1.1\textwidth, height=4.5cm]{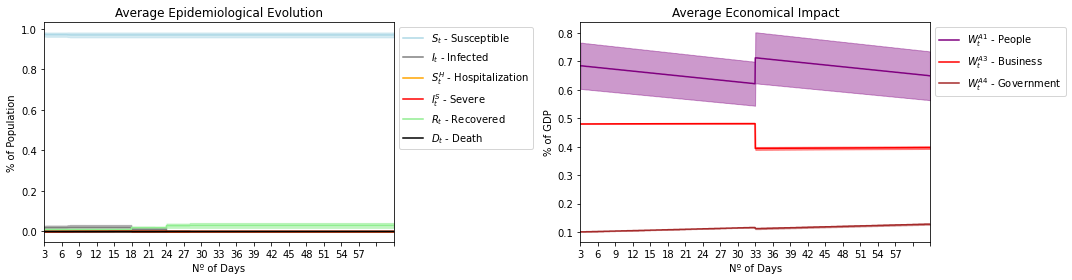}
    \caption{Daily averaged response variables for Scenario 2}
    \label{fig:scenario2}
\end{figure}

Considering the economic point of view, see Figure \ref{fig:scenario2}, this scenario is the worst for the industry because the $A1$ agents cannot generate wealth, but keep receiving their labor incomes\footnote{In our simulation, people can not get fired, which means that the onus of keeping them at home is for the company. In practice this may generate unemployment.}. $A3$ does not have income, but keeps paying taxes to $A4$ and labor expenses to $A1$. In this scenario, after two months, the businesses lost 20\% of its GDP share, see $W^{A3}_{S,T}$ in Figure \ref{fig:scenario2}.

The key point for the success of lockdown policy is staying at home (voluntarily or under laws). Economical countermeasures to its harm can also be adopted by $A4$, as tax exemptions and universal income, in order to minimize the wealth losses. In the impossibility of implementing this scenario, another one that considers protective and distance measures should be evaluated.

\subsection{Scenario 3: Conditional Lockdown}
\label{subsec:result_scenario_3}

This scenario imposes the same restrictions on $A1$ mobility presented in scenario 2, but conditionally. In the system, when the infection curve grows above a certain threshold, $I_t \geq 0.05$, the lockdown restrictions are activated, being released when $I_t \leq 0.05$.

\begin{figure}[h!]
    \centering
    \includegraphics[width=1.1\textwidth, height=4.5cm]{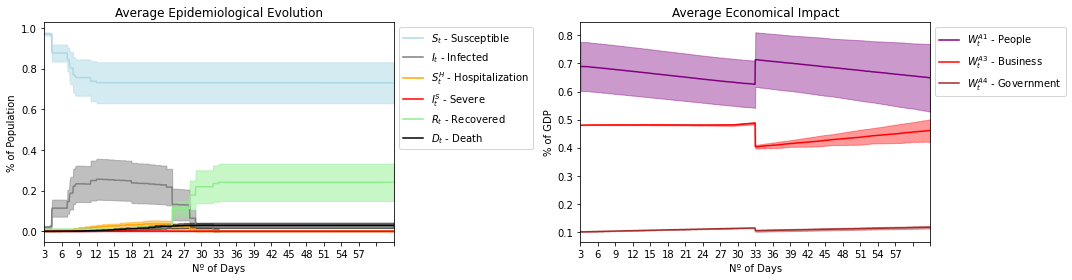}
    \caption{Daily averaged response variables for Scenario 3}
    \label{fig:scenario3}
\end{figure}

As we can see in Figure \ref{fig:scenario3}, the viral spreading represented by the infection curve $I_t$ is controlled, not allowing the explosion of $D_t$ curve. Economically, recession can be observed during the lockdown period, $W^{i}_{3,t}$ lower than $W^{i}_{B,t} \forall i$, but as soon as the restrictions are released the business performance is recovered. $W^{A3}_{3,t} $ remains below $W^{A3}_{B,t}$ but above the complete lockdown curve $W^{i}_{2,t}$.

Less conservative than scenario 2 (and also less efficient in terms of $D_t$), this scenario was implemented in New Zealand \cite{Cousins2020}, and it depends on an effective healthcare system that is capable of carrying out the necessary tests in the population, granting reliability in $I_t$ estimates and, as in scenario 2, the governmental ability to enforce the social isolation. 

\subsection{Scenario 4: Vertical Isolation}
\label{subsec:result_scenario_4}

Vertical isolation is the name given to the social intervention policy where the known infected people and the known risk groups -- elderly and people with pre-existent diseases -- are kept in social isolation, whereas young people and adults are allowed to work regularly. This policy has, for instance, been advocated by the Brazilian president\footnote{See \url{https://agenciabrasil.ebc.com.br/en/politica/noticia/2020-04/bolsonaro-brazil-must-not-be-informed-through-panic} - Acessed:  June 03, 2020, and \url{https://www.bbc.com/814portuguese/internacional-52043112} - Acessed:  June 03, 2020}. 


In terms of the proposed model, over $65$, below $18$ years old and symptomatic regardless of the age stay at home.

\begin{figure}[h!]
    \centering
    \includegraphics[width=1.1\textwidth, height=4.5cm]{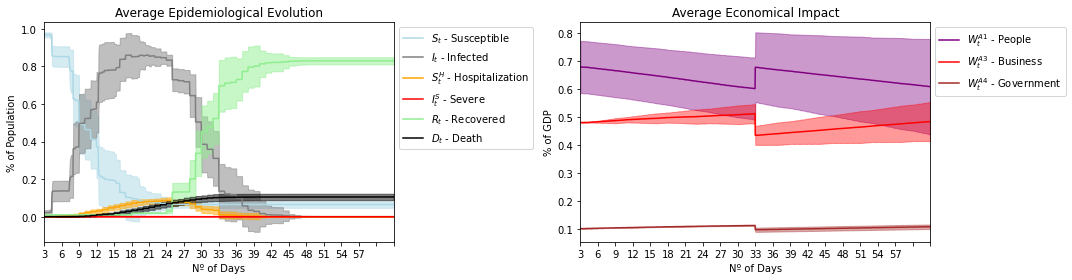}
    \caption{Daily averaged response variables for Scenario 4}
    \label{fig:scenario4}
\end{figure}

The assumption of this policy is that all the people outside the risk groups would not develop the severe cases of the disease. This assumption was proved to be fragile and this policy showed to be ineffective by \cite{duczmal2020_distanciamentovertical}. The results shown in Figure \ref{fig:scenario4} are in accordance with the literature  \cite{duczmal2020_distanciamentovertical} and produced almost the same epidemiological and economical results of Scenario 1, i.e., the same results of doing nothing.




\subsection{Scenario 5: Partial Isolation}
\label{subsec:result_scenario_5}

In the scenarios with lockdown (2 and 3), the mobility of all agents must be restricted, requiring restrictive public policies enforced by the government. When these policies are non-existent or are not taken seriously by the entire population, partial isolation levels are reached. The partial isolation level $IL \in [0,1]$ means the percentage of the population that is fulfilling the isolation, while the remaining $1 - IL$ is not.

Then, it is possible to define that in the lockdown $IL \geq 0.9$, considering that essential services and a few industries can not stop in order to avoid supply breakdown. On the other hand, the scenarios 0 and 1 have $IL \leq 0.1$, and the scenario 4 has $IL \approx 0.2$, because of the age distribution and the definition of risk groups.

This scenario aims to assess the effects of intermediate $IL$s. It was simulated by randomly choosing agents $A1$ with probability $IL \gets 0.5$ to stay at home.

\begin{figure}[h!]
    \centering
    \includegraphics[width=1.1\textwidth, height=4.5cm]{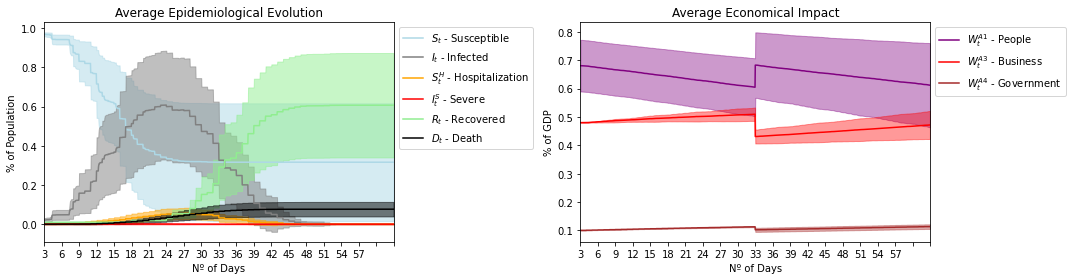}
    \caption{Daily averaged response variables for Scenario 5}
    \label{fig:scenario5}
\end{figure}

Observing the results in Figure \ref{fig:scenario5}, although the $I_t$ curve is flattened when compared with scenarios $B$ and 4, it is still less efficient than scenarios 2 and 3. Notice the $D_t$ still grows exponentially before reaching the peak. For the economic perspective, this scenario behaves similarly to the baseline. These metrics offer evidence that $IL \gets 0.5$ is not enough for effective epidemiological control, and a level of isolation greater than that is recommended.

\begin{figure}[h!]
    \centering
    \includegraphics[width=\textwidth, height=4.5cm]{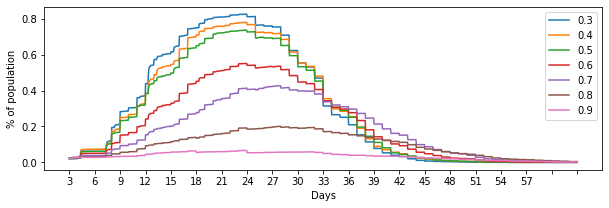}
    \caption{Infection curves by varying values of partial isolation level ($IL$).}
    \label{fig:isolation_levels}
\end{figure}

\begin{figure}[h!]
    \centering
    \includegraphics[width=\textwidth, height=4.5cm]{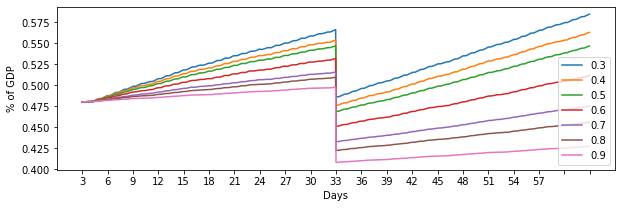}
    \caption{$W^{A3}_{S,t}$ curves by varying values of partial isolation level ($IL$).}
    \label{fig:gdp_isolation_levels}
\end{figure}

The impact of different isolation levels can be seen in Figures \ref{fig:isolation_levels} and \ref{fig:gdp_isolation_levels}, for $IL \in [0.3, 0.9]$, which represents the response of epidemiological and economical curves for increasing $IL$. In Figure \ref{fig:isolation_levels} it is possible to see how the infection curve $I_t$ flattens as the isolation level increases from no isolation towards  lockdown. Figure \ref{fig:gdp_isolation_levels} shows that as the value of $IL$ increases,  wealth loss of the A3 agents is higher, represented by $W^{A3}_{S,t}$ curve, showing the importance of agent's mobility in the economy. 


\subsection{Scenario 6: Use of Face Masks }
\label{subsec:result_scenario_6}



Evidence was found about the use of masks and gloves as measures against viral spreading \cite{chu2020physical}. This scenario represents the policy of mandatory usage of face masks and physical distancing, but without imposing restrictions on the mobility of agents. 

\begin{figure}[h!]
    \centering
    \includegraphics[width=1.1\textwidth, height=4.5cm]{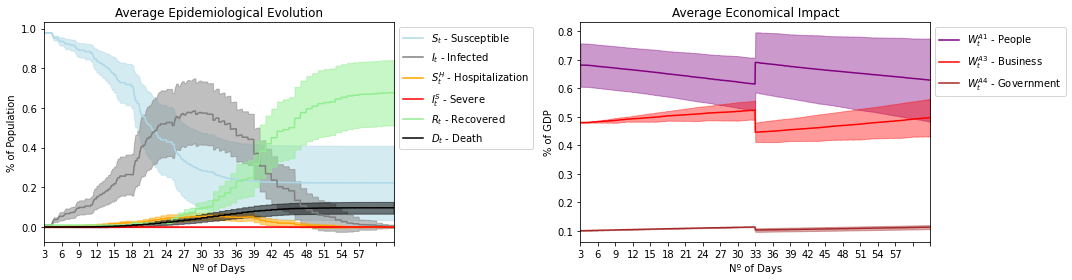}
    \caption{Daily averaged response variables for Scenario 6}
    \label{fig:scenario6}
\end{figure}

This scenario was implemented by reducing the contagion distance $\beta_1 = 0.5$ and the contagion rate $\beta_2 = 0.3$ as the effect of using masks and physical distancing. Figure \ref{fig:scenario6} shows a flatter $I_t$ curve when compared to scenario 5 while still keeping economic performance close to $B$. Notice, however, that $D_t$ is significantly higher when compared with scenarios 2 and 3. 

\subsection{Scenario 7: Use of Face Masks and 50\% of Social Isolation}
\label{subsec:result_scenario_7}

This scenario combines the policies used in the scenarios 5 and 6, granting the necessary use of face masks plus partial isolation of the population. This scenario was implemented by using $\beta_1 = 0.5$, $\beta_2 = 0.3$ and $IL = 0.5$.

\begin{figure}[h!]
    \centering
    \includegraphics[width=1.1\textwidth, height=4.5cm]{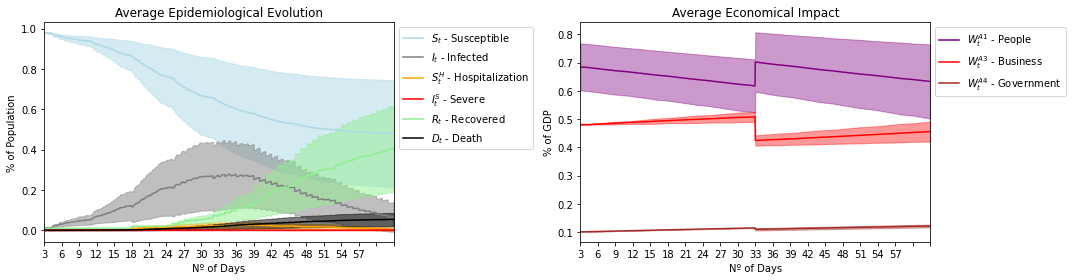}
    \caption{Daily averaged response variables for Scenario 7}
    \label{fig:scenario7}
\end{figure}

Figure \ref{fig:scenario7} shows the dynamics of this scenario. Although the $D_t$ is still above the values of scenarios 2 and 3, it presents less resistance from the general population. The $I_t$ is flattened, and the economy, despite the downturn, suffers less than it would in scenarios with lockdown. This scenario has already been discussed in \cite{chu2020physical} with similar results.

\subsection{Comparisons Among the Scenarios}

\begin{figure}[h!]
    \centering
    \includegraphics[width=\textwidth, height=5cm]{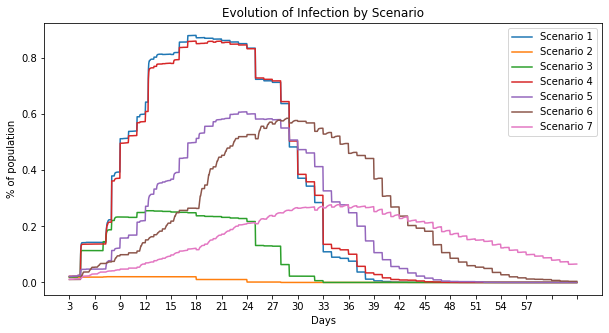}
    \caption{Infection evolution for the several scenarios}
    \label{fig:infections}
\end{figure}

\begin{figure}[h!]
    \centering
    \includegraphics[width=\textwidth, height=6cm]{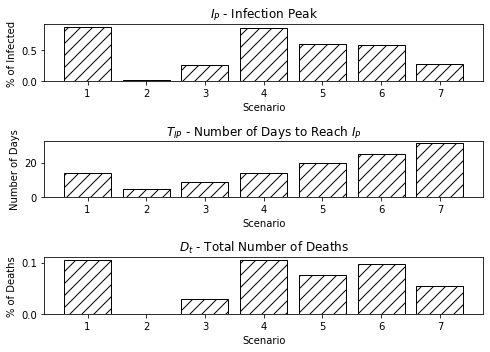}
    \caption{Death evolution for the several scenarios}
    \label{fig:epidemiological}
\end{figure}

The $I_t$ curves (averages) of each scenario are shown in Figure \ref{fig:infections}.  There, the effects of each intervention policy in flattening the curve can be observed and contrasted. The  epidemiological effectiveness of the scenarios are shown in Figure \ref{fig:epidemiological}, which compares the infection peak $I_P$ reached in each case, the number of days $T_{IP}$ to reach the peak $I_P$ and the max number of deaths $D_t$ (as a proportion of the population). 

As expected, scenarios 2 and 3 have the best epidemiological values followed by scenario 7. 

\begin{figure}[h!]
    \centering
    \includegraphics[width=\textwidth]{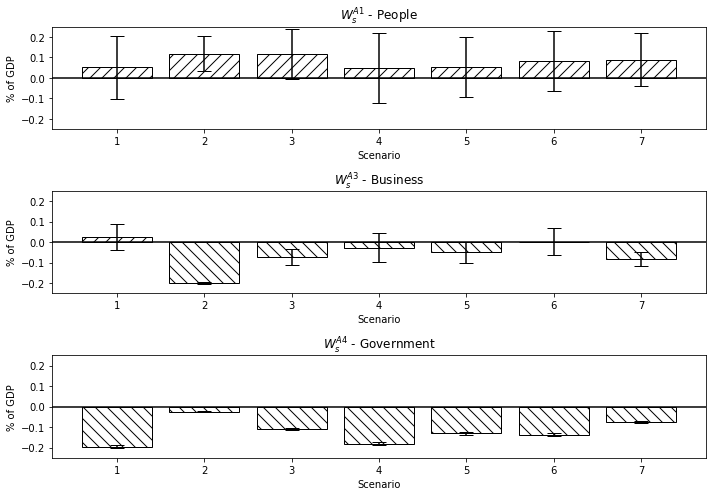}
    \caption{Economical result of each scenario  compared to Scenario 0 by response variable}
    \label{fig:economy}
\end{figure}

Figure \ref{fig:economy} shows the economic result of each scenario for the agent types $A1$, $A3$ and $A4$. Assuming that businesses are not firing anyone, from the point of view of the citizen, scenarios 2 and 3 are not economically damaging. On the other hand, the same scenarios are the worst from the business perspective. At this point, it is important to explain that the expenses of government in our simulation are related with the costs of the healthcare system. Thus, in scenarios with a high number of deaths, such as scenarios 1 and 4, the cost of maintaining the healthcare system is increased which demands an increase of public expenses. 

\begin{figure}[h!]
    \centering
    \includegraphics[width=1\textwidth, height=5cm]{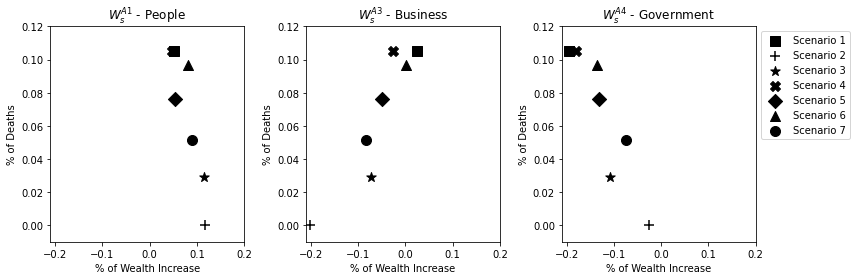}
    \caption{Percentage of deaths versus percentage of GDP variation}
    \label{fig:epidem_ecom}
\end{figure}

Figure \ref{fig:epidem_ecom} shows the scatter plots of the wealth increase (with respect to the baseline) of each type of agent by the percentage of deaths in the populations. It can be seen that, from a life preservation perspective, there is no better policy than the lockdown (scenario 2). Furthermore, in the simulated model, scenario 2 Pareto dominates\footnote{Given a set of criteria, \textit{Pareto optimality} can be defined as a situation where no individual criterion can be better off without making at least one other criterion worse off. Given an initial situation, a \textit{Pareto improvement} is a new situation where there will be gains in all criteria. A situation is called \textit{Pareto dominated} if it has a Pareto improvement. Finally, a situation is called \textit{Pareto optimal} if no change could lead to an improvement in all the objectives.} all the scenarios for both people an government. On the other hand, it represents the worst case, financially, for businesses.

In the impossibility of enforcing a lockdown (discarding scenarios 2 and 3), which may happen in underdeveloped countries, the best solution is represented by scenario 7. From the remaining Pareto optimal solutions for businesses, it is the one with lowest number of deaths. It also becomes the best solution for government and people in both wealth and number of deaths. 

\section{Conclusion}
\label{sec:conclusion}

The COVID-19 pandemic brought to humankind many challenges, including the demand for new medical treatments, social policies and economical approaches. The fast response of the scientific community to deal with coronavirus was divided into studies of the epidemiological aspects, proposals of new treatments and diagnostic tools and new models to forecast the viral spreading, including SIR and SEIR models among others. Nonetheless, few studies focused on looking at the pandemics as a governmental policy-making problem. With this viewpoint, although the epidemiological aspects are priority, the social and economical aspects can not be neglected.

The present work proposed an Agent-Based Model (ABM) that simulates the  epidemiological and economical effects of COVID-19 pandemic in a closed society, whose results can be generalized for wider contexts and used by governmental rulers to prospect social policies and assess its potential effectiveness in real scenarios.

The model was encapsulated in the free and open source software library COVID-ABS, which contains 29 epidemiological, social, demographic and economic input parameters, and 10 output response variables. New features can be designed and the library can be easily extended to other scenarios.

In a wider perspective, the proposed approach can be used as a decision-support system for the governments and scientific community. Policy-makers can design scenarios and evaluate the effectiveness of social interventions through different simulations, and analyse how the $P$ parameters, in the time horizon of $T$, can affect the response variables $\Theta_t$.

Seven different scenarios were elaborated to reflect specific social interventions. Lockdown and conditional lockdown were the best evaluated scenarios in preserving lives. These scenarios present a slower evolution of the epidemic, a smaller number of infections and deaths. Given the impossibility of implementing lockdown policies, the scenario with 50\%  of social isolation with using masks and physical distancing was the best approach in the preservation of lives. On the other hand, the vertical isolation scenario is totally ineffective and resembles the ``Do nothing'' scenario.


The results showed that COVID-ABS approach was capable to effectively simulate social intervention scenarios in line with the results presented in the literature. Also, the results showed that policies adopted by some countries, for instance US, Sweden and Brazil, are ineffective when the objective is to preserve lives. Governments that chose to preserve the economy by not using severe isolation policies, fatally reached a situation with a high cost in human lives, and still embittered economic losses. The evidence provided by the simulation model shows that there is a false dichotomy between healthcare and the economy. In the scenarios where it was tried to save the economy by not taking hard social isolation policies, consequently, the social costs ended up impacting negatively into the economy.

COVID-ABS is an open software and can be easily extended and customized. Also, new scenarios can be designed, taking into consideration the specificities of each region under study.  
Future research aims to improve the model by implementing mechanisms to close and open companies as well as allowing people to get fired. In addition, it will be integrated with optimization libraries, for  automatic scenario creation, and multi-criteria decision making tools that could help governmental crisis committees to plan and manage the social policies to mitigate the COVID-19 effects.

\section*{Acknowledgements}
Petr\^onio Silva, Paulo Batista and Helder Seixas would like to thank the financial support given by the Instituto Federal do Norte de Minas Gerais, Brazil.

Marcos A. Alves declares that this work has been supported by the Brazilian agency CAPES. 

Frederico Gadelha Guimar\~aes would like to thank the support given by the Brazilian Agencies CNPq (grant no. 306850/2016-8) and FAPEMIG.






\bibliographystyle{auxiliary/elsarticle-num-names.bst}
\bibliography{bibliography.bib}







\end{document}